\documentclass[sigconf]{acmart}
\renewcommand{\footnotetextcopyrightpermission}[1]{}
\AtBeginDocument{%
  }

\usepackage{listings}
\usepackage{xcolor}
\usepackage{tabularx}
\usepackage{float}
\PassOptionsToPackage{table}{xcolor}
\usepackage{makecell}
\usepackage{booktabs}  
\usepackage[most]{tcolorbox}
\usepackage{needspace}

\acmConference[FinIR '25]{FinIR: The 2nd Workshop on Financial Information Retrieval in the Era of Generative AI}{July 17, 2025}{Padua, Italy}

\begin{document}

\title{
Agentic Retrieval of Topics and Insights from Earnings Calls
}

\author{Anant Gupta}\author{Rajarshi Bhowmik}\author{Geoffrey Gunow}
\affiliation{%
  \institution{Bloomberg}  
  \country{USA}
}

\renewcommand{\shortauthors}{Gupta et al.}

\begin{abstract}
  Tracking the strategic focus of companies through topics in their earnings calls is a key task in financial analysis. However, as industries evolve, traditional topic modeling techniques struggle to dynamically capture emerging topics and their relationships. In this work, we propose an LLM-agent driven approach to discover and retrieve emerging topics from quarterly earnings calls. We propose an LLM-agent to extract topics from documents, structure them into a hierarchical ontology, and establish relationships between new and existing topics through a topic ontology. We demonstrate the use of extracted topics to infer company-level insights and emerging trends over time. We evaluate our approach by measuring ontology coherence, topic evolution accuracy, and its ability to surface emerging financial trends.
\end{abstract}





\maketitle

\section{Introduction}
In the domain of finance, earnings calls—quarterly discussions between company executives and financial analysts—are essential events where companies disclose performance results, strategic initiatives, and address analyst inquiries. These calls provide critical insights into company operations, research and development trajectories, and broader industry trends, making them valuable resources for financial analysis \cite{arslan2016managers}.

Financial analysts closely examine these earnings calls to assess corporate financial health, strategic direction, and evolving market positions. Given the substantial volume and complexity of these calls, AI-driven solutions have emerged to assist analysts with various tasks, such as summarization \cite{yang2024evaluating}, sentiment analysis \cite{hajek2013evaluating, araci2019finbert, wu2024bloomberggpt}, and topic extraction \cite{zhao2021topic}. While summarization and question-answering \cite{wu2024bloomberggpt} typically focuses on an individual or a small set of documents, topic extraction can scale to larger corpora but often requires a pre-vetted list of financially relevant topics to be detected, limiting their ability to dynamically identify new and emerging themes. A pre-vetted static list of topics ensures quality control, aligns with analyst expectations, and filters meaningful financial signal from noise, making the output more actionable and consistent for downstream analysis.

In response to this challenge, our work proposes an LLM-agent driven solution specifically designed to dynamically track and analyze financially-relevant topics across large corpora of earnings calls. We break this complex problem into two core subproblems:

\textit{1.} Identifying relevant and emerging topics is essential yet challenging. 
Traditional supervised and unsupervised topic discovery models fall short on discovering emerging topics due to their inherent limitations. \textit{1.a.} Supervised methods are restricted to a predefined set of topics and needs substantial amount of labeled data for model training \cite{gandouet2024distilled}. Consequently, they frequently miss crucial emergent topics— such as sudden macroeconomic shifts like the onset of COVID-19 — that are critically relevant yet lack historical data. \textit{1.b.} The unsupervised topic models, on the other hand, do not capture the contextual relationships between words or phrases within a topic, which can limit their ability to capture nuanced meanings. The topics generated by unsupervised models are often represented as lists of words, agnostic to financial-relevance, making it difficult to assign meaningful, well-defined labels without human intervention. Additionally, traditional topic models such as Latent Dirichlet Allocation (LDA) \cite{blei2003latent} requires a collection of documents (also known as corpus) to identify topics. Identifying topics in each document independently is challenging for such models. To this end, we propose an LLM-agent to discover topics in a document.

\textit{2.} Effective management of topic knowledge requires structuring topics within an ontology that captures hierarchical relationships and interconnections \cite{aggarwal2024large}. Topics rarely exist in isolation; they often manifest as interconnected parent and child concepts. Thus, a topic ontology provides critical context of topic relationships, significantly enhancing an analyst's comprehension of topic evolution.

Addressing these two problems, we introduce a novel LLM-powered agentic framework. This autonomous decision-making system uses Large Language Models (LLMs) to iteratively extract emerging topics, assess their novelty, and systematically integrate them into a continuously evolving topic ontology. Our methodology facilitates the nuanced tracking of evolving financial narratives, offering analysts comprehensive macro-level insights into company and industry dynamics over time.

Specifically, our framework enables users to identify topics discussed across a series of earnings call documents, track how the narrative surrounding these topics evolves over time, and compare these narratives between companies, or sectors \cite{gandouet2024distilled}. 

In this paper, we demonstrate the capabilities of the proposed framework by discovering and constructing a topic ontology over a well-scoped out dataset. We compare the performance of different components of this framework - quantitatively and qualitatively. We use the discovered topics to extract macro-level insights over companies and their competitors over a period of time. In the end, we note certain limitations such as the lack of labeled data, the subjective nature of the task, and suggest future directions of this work.



\begin{table*}[t]
\centering
\small
\begin{tabularx}{\textwidth}{p{2cm}X}
\toprule
\textbf{Full \newline Self-Driving (FSD)} & 
\textbf{Text}: I mean, just to give you one example, so again, there's a biased example, but I have a more than 20-mile commute into the factory almost every day. And I have zero interventions on the latest stack, and the car just literatized me a lot. And especially with the latest version, wherein we're also tracking your eye movement, the steering wheel lag is almost not there, as long as you're not wearing sunglasses.
\vspace{0.5em}
\newline
\textbf{\textcolor{blue}{Excerpt}}: \textcolor{blue}{More than 20-mile commute with zero interventions on the latest stack, with the car literalizing the driver, Latest version tracks eye movement, reducing steering wheel lag when not wearing sunglasses} \\
\midrule
\textbf{Cost Reduction} & 
\textbf{Text}: So, its in our control. We have to take cost out. We know that. That's what we're marching towards. And we're understanding the dynamics and the competitiveness of the market equation as we set up the cost structure for our second-gen vehicles. And as Jim said, we pushed out the three-row SUV because we need more cost to come out of that for that to be at the margin levels we expect.
\vspace{0.5em}
\newline
\textbf{\textcolor{blue}{Excerpt}}: \textcolor{blue}{Company needs to reduce costs, particularly for second-generation vehicles, Three-row SUV launch delayed to achieve desired margin levels through cost reductions} \\
\addlinespace
\midrule
\textbf{Enterprise Adoption} & 
\textbf{Text}: We've recently introduced two new platforms. The 8cx Gen 3 and 7c+- Gen 3. In [ph]ex CES, we highlighted broad support from ecosystem partners, including Acer, ASUS, HP, Lenovo, and Microsoft. As well as 200 enterprise customers currently testing or deploying Windows on Snapdragon laptops in two-in-one devices.
\vspace{0.5em}
\newline
\textbf{\textcolor{blue}{Excerpt}}: \textcolor{blue}{200 enterprise customers are testing or deploying Windows on Snapdragon laptops and two-in-one devices} \\
\addlinespace
\midrule
\textbf{Data Center \newline Demand} & 
\textbf{Text}: As my real question, can you speak to your data center visibility into 2022 and beyond? And within this outlook, can you talk to traditional cloud versus industry verticals and then perhaps emerging opportunities like Omniverse and others? Would love to get a sense of kind of what you're seeing today. And then as part of that how you're planning to secure foundry and other supply to support that growth? Thank you.
\vspace{0.5em}
\newline
\textbf{\textcolor{blue}{Excerpt}}: \textcolor{blue}{Question about data center visibility into 2022 and beyond, including traditional cloud and industry verticals} \\
\addlinespace
\bottomrule
\end{tabularx}
\caption{Examples of discovered financial topics and generated excerpts from earnings calls text. Note that while the topic may not appear verbatim, it is semantically represented throughout the excerpted text.}
\label{tab:topic_examples}
\end{table*}

\section{Dataset}

The dataset selected for this study is carefully scoped to illustrate how insights can be effectively extracted at the company or sector level, aiding financial analysts in gaining a deeper understanding of specific market spaces. Our dataset comprises earnings call documents from two distinct industry sectors, with each sector represented by six companies considered direct competitors based on their product offerings. Table \ref{tab:sector_companies} shows the companies selected for this study. We choose \textit{EVs} and \textit{Semiconductors} as our sectors which are often discussed in mainstream media and would be easier for the readers who may not be subject matter experts to understand the insights and relate to them.

\begin{figure}[htbp]
    \centering
    \includegraphics[width=\linewidth, alt={Paragraph lengths}]{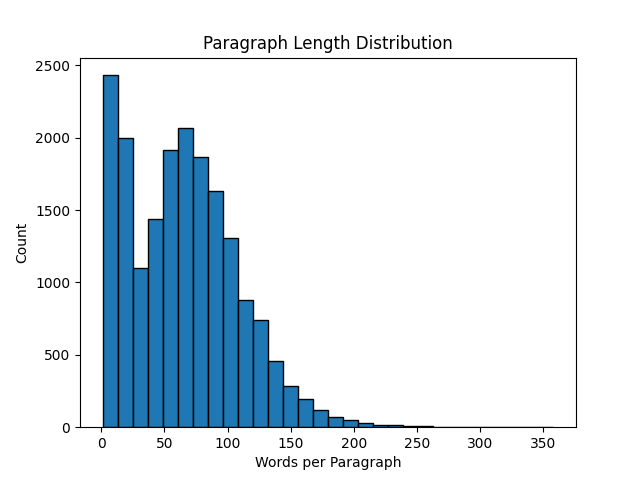}
    \caption{Distribution of paragraph lengths (defined by number of words).}
    \label{fig:paragraphs}
\end{figure}

We include earnings call documents over a 3-year time period from late 2021 to mid-late 2024 for each company in our dataset. The textual data within these documents is segmented by paragraphs, approximately aligning with the inherent structure of earnings calls: typically, such calls start with commentary provided by various company executives, followed by a Q\&A segment with financial analysts. Although commentary occasionally spans multiple paragraphs, the paragraph-level segmentation provides a practical and representative granularity for individual insights extraction, aligning well with the data storage and processing methodologies employed in this research. Table \ref{tab:dataset_stats} shows statistics over this dataset and Figure \ref{fig:paragraphs} shows frequency distribution of the paragraphs.

\begin{table}[h!]
\centering
\begin{tabular}{|p{2cm}|p{5.5cm}|}
\hline
\textbf{Sector} & \textbf{Companies (and ticker symbols)} \\
\hline
Electric \newline Vehicles & 
Tesla (TSLA) \newline Ford Motor Company (F) \newline General Motors (GM)\newline Rivian Automotive (RIVN) \newline Lucid Group (LCID) \newline Polestar Automotive (PSNY) \\
\hline
Semiconductors & 
NVIDIA Corporation (NVDA) \newline Advanced Micro Devices (AMD) \newline Intel Corporation (INTC) \newline ASML Holding (ASML) \newline Broadcom Inc. (AVGO) \newline Qualcomm Incorporated (QCOM) \\
\hline
\end{tabular}
\caption{Companies and sectors in the dataset.}
\label{tab:sector_companies}
\end{table}

\begin{table}[h!]
\centering
\begin{tabular}{|l|l|}
\hline
\textbf{Statistic} & \textbf{Count} \\
\hline
Total transcripts (documents) &  141 \\
Total paragraphs & 18637 \\
Total quarters covered & 12 \\
Vocabulary size (unique words) & 15014 \\
Average paragraph length (words) & 64.79 ± 43.71 \\
Average document length (words) & 8563.57 \\
Average document length (paragraphs) & 132.18 \\
Average sentence length (words) & 16.88 \\
\hline
\end{tabular}
\caption{Statistics of the earnings calls dataset.}
\label{tab:dataset_stats}
\end{table}

\section{Methodology}

\subsection{Agentic System}
\begin{figure}[htbp]
    \centering
    \includegraphics[width=\linewidth, alt={agentic system}]{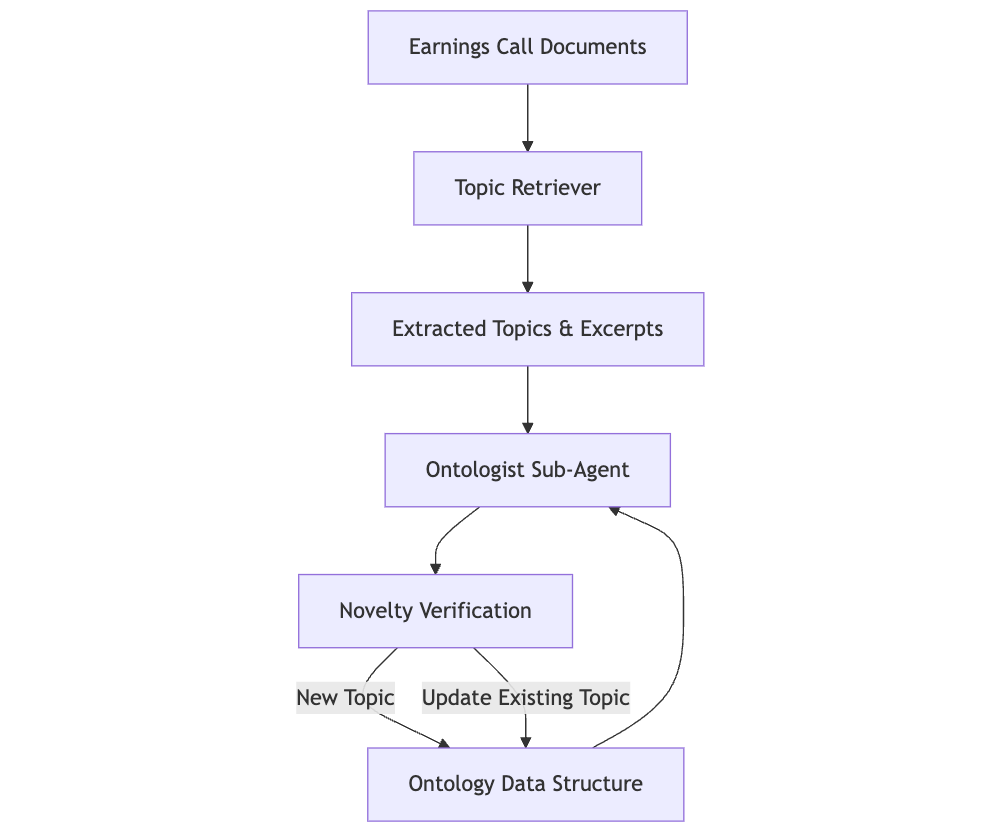}
    \caption{Overview of the agentic system.}
    \label{fig:overview}
\end{figure}

Our proposed methodology features a LLM-driven agentic system that autonomously retrieves and organizes topics from large corpora of earnings call documents into a continuously evolving ontology. This agentic framework comprises three interrelated components: a topic retriever, an ontology data structure, and an ontologist agent responsible for topic validation and integration. See Figure \ref{fig:overview} for an overview of the system.

Initially, the topic retriever analyzes paragraph content to identify and extract financially-relevant topics along with corresponding textual excerpts. The ontologist then assesses each identified topic's novelty by verifying its presence within the existing ontology, systematically adding new topics or updating existing ones to reflect ongoing narrative developments. The ontology stores these topics by maintaining hierarchical relationships and contextual associations among them. Detailed descriptions of each module follow in subsequent sections.


\subsection{Topic Retriever}

The Topic Retriever is a core component of our LLM-driven agentic framework, designed specifically to extract pertinent topics from earnings call documents over a span of text. This module uses a carefully designed prompt structure (Appendix \ref{appendix:prompt_topic}) utilizing LLMs to systematically identify multiple relevant financial topics within textual data.

In our system, a "topic" is formally defined as a distinct conceptual entity pertinent to financial analysts, ranging from general financial themes such as "guidance", "capital expenditures (capex)", "dividends", and "macroeconomic conditions", to more specific industry terminologies and initiatives such as "Full self-driving," "Data Center", and "Regulatory Developments". The Topic Retriever is explicitly instructed to extract topics individually without concatenating distinct concepts.

Alongside identifying topics, the Retriever also generates succinct excerpts for each topic mentioned. These excerpts are condensed, contextual summaries that capture the essence of the topic's mention within the broader text. Extracting excerpts addresses scenarios where topics might be recurrently discussed or embedded within larger commentaries, thereby providing analysts with a readily accessible and contextually meaningful snapshot while exploring topic insights. Table \ref{tab:topic_examples} shows examples of some of the extracted topics, original text, and generated excerpt.

The retrieved topic information is used to construct the topic ontology. The topic identifier from the ontology and the excerpt are stored along with the original text as enrichment.

\subsection{Topics Ontology}

The Topics Ontology is structured as a tree, where each node represents a distinct topic. Nodes in this ontology follow a hierarchical parent-child relationship, where child nodes represent subcategories of their respective parent topics. In this work, we maintain the topic topology in a simple tree structure. However, it can  easily be extended to other topological structures (e.g., Directed Acyclic Graph) for building and maintaining more complex topic relationships. Table \ref{tab:node_properties} outlines the properties encapsulated within each node.

Constructing such an ontology comes with the challenge of clearly defining the parent-child relationships among topics, which is nontrivial due to potential overlap and semantic ambiguity. We let the ontologist agent guide this relationship among nodes with the use of LLMs.


\begin{table}[h!]
\centering
\begin{tabular}{|l|l|}
\hline
\textbf{Property} & \textbf{Type} \\
\hline
Topic ID &  \textit{UUID} \\
Topic Name & \textit{string} \\
Topic Aliases & \textit{List[string]} \\
Created on &  \textit{datetime} \\
Updated on & \textit{datetime} \\
\hline
\end{tabular}
\caption{Properties of a node in Topics Ontology.}
\label{tab:node_properties}
\end{table}

\subsection{Ontologist Agent}

The Ontologist is a critical LLM-based agent responsible for constructing and maintaining the topic ontology. The operational workflow of this agent, detailed in Figure \ref{fig:overview}, encompasses two primary operations: topic existence and topic insertion.

\subsubsection{Topic Existence}

When the Topic Retriever extracts topics from textual data, semantic consistency must be ensured. Due to the inherent variability in topic naming (e.g., \textit{"M\&A"} versus \textit{"Mergers \& Acquisitions"}), simple equality checks are insufficient. Thus, we utilize semantic similarity assessments powered by LLMs to evaluate if a newly retrieved topic aligns with existing ontology nodes or their aliases. When a semantic match is identified, the newly retrieved topic name is appended as an alias to the existing topic node in the ontology. This approach not only optimizes topic retrieval through semantic matching but may also facilitate other downstream retrieval tasks by recognizing synonymous terms.

A carefully structured semantic similarity prompt (Appendix \ref{appendix:prompt_topic_exists}) guides this verification process, ensuring precise semantic equivalence rather than broader categorical matches, through detailed criteria, considering scope, specificity, domain context, and relational equivalence.

\subsubsection{Topic Insertion}

If no semantic equivalence is identified, the agent determines the most suitable position within the ontology for the new topic. This insertion process involves an LLM-driven analysis to identify a potential parent topic node, starting from top-down to first identify a broader topic node and iteratively seek its children until the most granular parent for that topic is found. If an appropriate parent is found, the new topic node is integrated accordingly. Otherwise, the topic is introduced as an isolated node at the ontology's root level.

Both topic verification and insertion operations return unique topic IDs, enriching the textual data with structured identifiers and corresponding excerpts. This structured enrichment significantly enhances subsequent analytical capabilities.

\begin{figure}[htbp]
    \centering
    \includegraphics[width=\linewidth, alt={Smoothed trend of new topics discovered over time.}]{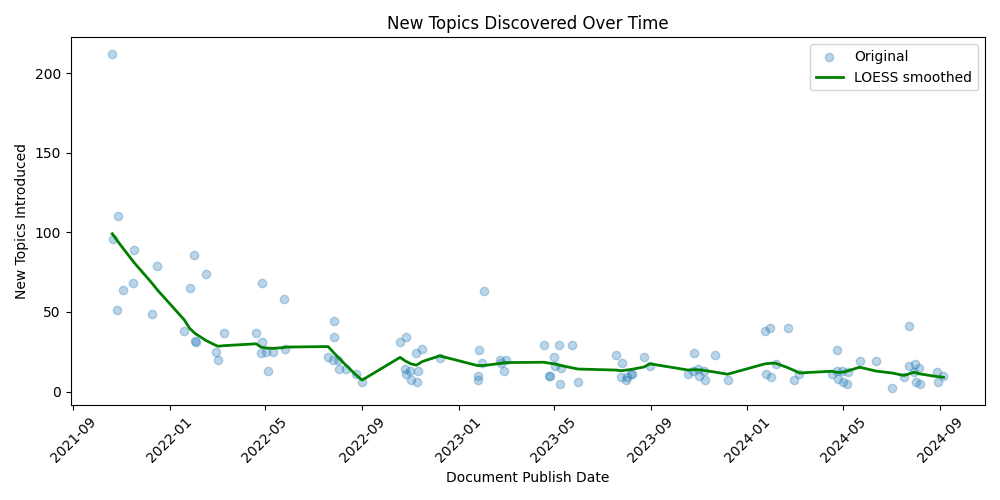}
    \caption{Count of topics discovered over time as we only insert topics that don't exist in the ontology. We apply \textit{Locally Estimated Scatterplot Smoothing (LOESS)} smoothing to highlight this trend.}
    \label{fig:new_topics}
\end{figure}


\section{Experiments}

\subsection{Topic Retrieval}
We benchmark our proposed topic retriever method against a well-established unsupervised baseline: Latent Dirichlet Allocation (LDA). In the absence of ground-truth topic labels, this method serves as a suitable baseline for qualitative evaluation. We adapt the open-source workflow from \cite{jansen2020machine} which trains a LDA model on earnings calls transcripts and use the same preprocessing and vectorizing methods mentioned in that work on our entire dataset. We train to identify 15 topics represented by 10 coherent keywords with 25 passes over the corpus.

The results from the model were observed to be uneven and sometimes not directly related financial topics. The heat map in Figure \ref{fig:lda_heatmap} (Appendix A.1) shows that topics T4 (keywords: \textit{cloud, service, platform, technology, solution}) and T11 (keywords: \textit{loan, capital, bank, credit, asset, fee}) contain semantically consistent, finance-relevant terms, whereas the remainder mix generic fillers such as \textit{thing, maybe, chief, officer} with domain words, making the topic boundaries hard to interpret. This observation is corroborated by the \textit{u\_mass} coherence scores: most topics fall between -0.4 and -2.5, with only three exceeding -0.8. 

We also conducted a supplementary study to compare the number of topics identified over time to contrast with Figure \ref{fig:new_topics} but we did not obtain any further meaningful results on stratifying the dataset further by time-period. The quantitative and qualitative signals suggest that standard LDA model struggles to produce well-formed “financial” topics on this corpus, reinforcing our preference for the proposed LLM-agentic retriever, which consistently yields coherent, hierarchically grounded topics in downstream analyses.

\subsection{Topic Ontology} Topic Ontology is constructed by iterating over the entire dataset. On each iteration of the paragraph data sample, the topic ontology is updated either with a new topic or a topic alias to an existing topic. To establish initial standardization of the topics at the top-level, we construct the ontology with a fixed set of seed topics at the root level (32) and the level below it (361). There are no constraints on where the new topics can be added in the ontology. Table \ref{tab:ontology-stats} shows the statistics of the final ontology after iterating over the entire dataset.

\subsubsection{Ontology Coherence Evaluation}
To evaluate the semantic coherence of our topic ontology in the absence of gold-standard labels, we conducted an embedding-based experiment using the \texttt{all-MiniLM-L6-v2} model from Sentence-Transformers \cite{reimers2019sentence}. We sampled a set of parent topics from the ontology and, for each, selected at most five associated child topics (ensuring each parent had at least two children). We computed the cosine similarity between the parent topic and each of its children, then averaged these scores to obtain a coherence score per parent node. As a baseline, we randomly sampled an unrelated parent topic and measured its similarity to the original set of child topics. Our results in Table \ref{tab:ontology-coherence} show that the true parent-child pairs exhibit substantially higher semantic similarity (average cosine similarity: 0.383) compared to random parent-child pairings (0.153), validating the structural integrity and conceptual closeness of our automatically generated topic hierarchy. This indicates that the GenAI-driven ontology construction process effectively groups semantically related financial topics under appropriate conceptual categories.

\begin{table}[ht]
\centering
\begin{tabular}{lr}
\toprule
\textbf{Metric} & \textbf{Value} \\
\midrule
Total nodes &  \texttt{3200} \\
Number of levels &  \texttt{4} \\
Number of leaf nodes &  \texttt{2986} \\
Average children per node &  \texttt{1 ± 9.62} \\
Average aliases per node &  \texttt{2.5 ± 5.56} \\
\midrule
\multicolumn{2}{c}{\textbf{Nodes per level}} \\
\midrule
Level 0 (Root) & \texttt{42} \\
Level 1 & \texttt{1249} \\
Level 2 & \texttt{1896} \\
Level 3 & \texttt{13} \\
\bottomrule
\end{tabular}
\caption{Structural statistics of the topic ontology tree.}
\label{tab:ontology-stats}
\end{table}

\section{Results: Topic Insights}
We demonstrate the value of the retrieved topics from earnings calls by extracting macro-level insights about companies and broader industry sectors. These insights are valuable to equity analysts who evaluate the performance of companies and sectors to understand their future direction and henceforth make investment decisions. The insights presented here represent a subset of the potential findings our method can yield. We aim to detail such other insights in future work. For the following analyses, we consider topics that aren't \textit{products} (e.g., H100, Cybertruck) as the mentions of product-topics are a function of the launch timeline wherein they start trending up closer to their launch and trending down when newer products are being launched.

\subsection{Trend Analysis}
Identifying emerging and declining trends within industries and companies is essential for financial analysts to predict future directions and make informed investment decisions. Detecting these shifts early can provide significant competitive advantages by enabling timely actions.

Here we define trends as \textit{topics exhibiting statistically significant changes in frequency of mentions over consecutive earnings calls}. We quantify the significance of trends using \textit{Kendall's tau test}, a non-parametric measure suitable for assessing the ordinal association and monotonic relationships within time-series data. \textit{Kendall's tau} is particularly relevant here because it effectively captures gradual shifts and persistent directional trends in topic mentions, making it robust against outliers and small sample sizes typical in quarterly financial reporting \cite{chen2022practical}.

\begin{figure*}[htbp]
    \centering
    \includegraphics[width=\textwidth]{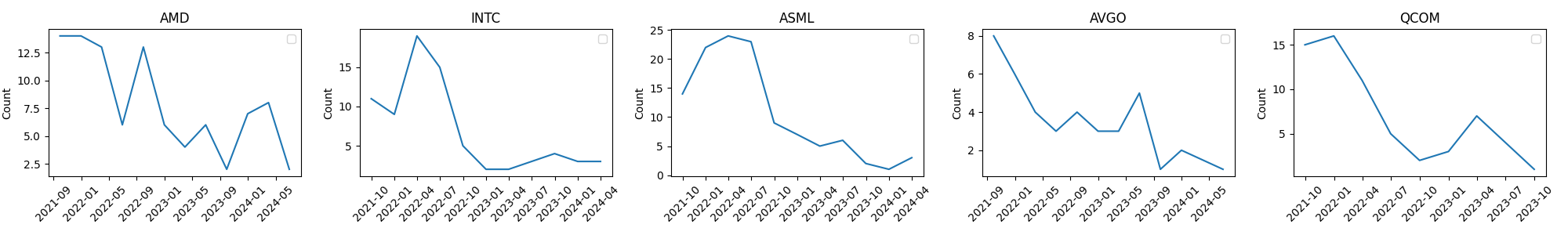}
    \caption{Trend of "Supply Chain" in semiconductor industry.}
    \label{fig:supply_chain_semi}
\end{figure*}

\begin{figure*}[htbp]
    \centering
    \includegraphics[width=\textwidth]{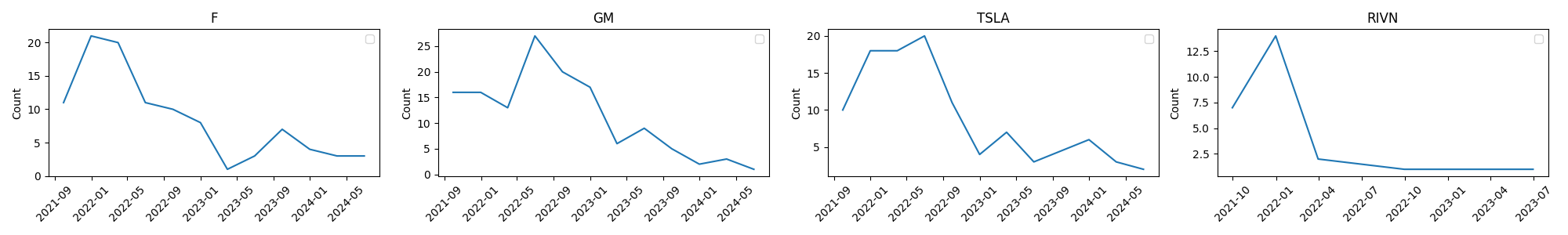}
    \caption{Trend of "Supply Chain" in EV industry.}
    \label{fig:supply_chain_ev}
\end{figure*}

Table \ref{tab:trending_down} and Table \ref{tab:trending_up} show the topics for each company that have been considered significant trends going in a particular direction. For example, we observe that the topic \textit{"Supply Chain"} in the semiconductor (Fig. \ref{fig:supply_chain_semi}) and EV sectors (Fig. \ref{fig:supply_chain_ev}) are being mentioned less over time, due to the decreasing challenges around that topic. We confirm this trend with Deloitte’s global semiconductor outlook 2025, which notes that “semiconductor supply chains worked well in 2024” even amid strong industry growth, and there was “no reason to believe 2025 supply chains will be less resilient” 
 \textcolor{blue}{\textit{(\href{https://www2.deloitte.com/us/en/insights/industry/technology/technology-media-telecom-outlooks/semiconductor-industry-outlook.html}{link})}}. We also found a similar report about supply chain trend in the EV sector where S\&P Global's research report notes: "With supply chain issues approaching a more normalized situation in 2023.." \textcolor{blue}{\textit{(\href{https://www.spglobal.com/mobility/en/research-analysis/the-semiconductor-shortage-is-mostly-over-for-the-auto-industry.html}{link})}}.
 
The practical value of our methodology lies primarily in its timeliness. By continuously integrating new earnings call data into our hierarchical topic ontology, we detect significant changes in topic relevance immediately upon document release. We further validate the advantage of our system by comparing our trend identification timelines against publicly available financial news and research sources. Our findings consistently demonstrate that the insights generated by our approach appear significantly earlier—often immediately after an earnings call—whereas comparable information typically emerges later in mainstream financial media.

\begin{table}[H]
\centering
\begin{tabular}{l|cccccc}
\hline
        & NVDA & AMD & INTC & ASML & AVGO & QCOM \\
\hline
NVDA    & 1.00 & \textbf{0.25} & 0.21 & 0.14 & 0.21 & 0.19 \\
AMD     & 0.25 & 1.00 & \textbf{0.34} & 0.15 & 0.23 & 0.26 \\
INTC    & 0.21 & \textbf{0.34} & 1.00 & 0.22 & 0.27 & 0.25 \\
ASML    & 0.14 & 0.15 & 0.22 & 1.00 & \textbf{0.18} & 0.14 \\
AVGO    & 0.21 & 0.23 & \textbf{0.27} & 0.18 & 1.00 & 0.18 \\
QCOM    & 0.19 & \textbf{0.26} & 0.25 & 0.14 & 0.18 & 1.00 \\
\hline
\end{tabular}
\caption{Jaccard Similarity of top 100 topics for Semiconductors.}
\label{tab:jaccard_similarity}
\end{table}

\begin{table}[H]
\centering
\begin{tabular}{l|cccccc}
\hline
        & TSLA & F & GM & RIVN & LCID & PSNY \\
\hline
TSLA    & 1.00 & 0.30 & \textbf{0.38} & 0.32 & 0.27 & 0.18 \\
F       & 0.30 & 1.00 & \textbf{0.50} & 0.36 & 0.31 & 0.26 \\
GM      & 0.38 & \textbf{0.50} & 1.00 & 0.35 & 0.32 & 0.26 \\
RIVN    & 0.32 & 0.36 & 0.35 & 1.00 & \textbf{0.42} & 0.28 \\
LCID    & 0.27 & 0.31 & 0.32 & \textbf{0.42} & 1.00 & 0.32 \\
PSNY    & 0.18 & 0.26 & 0.26 & 0.28 & \textbf{0.32} & 1.00 \\
\hline
\end{tabular}
\caption{Jaccard Similarity of top 100 topics for EVs.}
\label{tab:jaccard_similarity_ev}
\end{table}

\begin{table*}[ht]
\centering
\begin{tabular}{|l|c|l|c|}
\hline
\textbf{Parent Topic} & \textbf{Avg. Cosine Similarity} & \textbf{Random Topic} & \textbf{Avg. Cosine Similarity} \\
\hline
Retail and Consumer       & 0.376 & Country-Specific Markets        & 0.246 \\
Consumer Behavior         & 0.350 & Healthcare and Pharmaceuticals  & 0.121 \\
Forward-looking Guidance  & 0.236 & Industrial and Materials        & 0.084 \\
AI \& Machine Learning    & 0.483 & Batteries                       & 0.137 \\
Software                  & 0.523 & Operational Metrics             & 0.225 \\
M\&A                      & 0.193 & Generative AI                   & 0.089 \\
Margins                   & 0.620 & Industry-Specific Markets       & 0.151 \\
Generative AI             & 0.342 & Sensor Technology               & 0.088 \\
Metaverse                 & 0.195 & Cost Management                 & 0.145 \\
Business Transformation   & 0.512 & Economic Indicators             & 0.243 \\
\hline
\textbf{Average}          & \textbf{0.383} &                             & \textbf{0.153} \\
\hline
\end{tabular}
\caption{Comparison of average cosine similarity between actual parent-child topic pairs and randomly sampled parent for the same child pairs. Higher similarity values indicate better semantic coherence within the ontology structure.}
\label{tab:ontology-coherence}
\end{table*}

\begin{table}[ht]
\centering
\begin{tabular}{|p{2cm}|p{5.5cm}|}
\hline
\textbf{Company} & \textbf{Topics: trending down} \\
\hline
\multicolumn{2}{|c|}{\textbf{Industry: Semiconductors}} \\
\hline
QCOM & Supply Chain, consumer demand, M\&A, handset revenue \\
AVGO & Supply Chain, hyperscale, wireless revenue \\
INTC & Supply Chain, free cash flow, enterprise demand \\
AMD & Supply chain, market share \\
ASML & Supply chain, semiconductor demand \\
\hline
\multicolumn{2}{|c|}{\textbf{Industry: EVs}} \\
\hline
TSLA & Supply Chain, cost reduction, manufacturing, manufacturing ramp, supply Chain management \\
RIVN & Supply chain disruptions \\
F & Supply chain, EV production, inflation, business model, business restructring \\
PSNY & Macro environments \\
GM & Supply chain, chip shortage, material costs \\
\hline
\end{tabular}
\caption{Topics that are trending down over the time period considered. Refer figures \ref{fig:supply_chain_ev} and \ref{fig:supply_chain_semi} for "Supply Chain" plots.}
\label{tab:trending_down}
\end{table}

\begin{table}[ht]
\centering
\begin{tabular}{|p{2cm}|p{5.5cm}|}
\hline
\textbf{Company} & \textbf{Topics: trending up} \\
\hline
\multicolumn{2}{|c|}{\textbf{Industry: Semiconductors}} \\
\hline
NVDA & AI inference, AI Workloads transitions, generative AI \\
QCOM & AI PC, automotive Industry, on-device AI \\
INTC & Generative AI, operational metrics, wafer capacity \\
\hline
\multicolumn{2}{|c|}{\textbf{Industry: EVs}} \\
\hline
LCID & Batteries, cost reduction, software development \\
F & Product quality management, product launch timeline \\
PSNY & Working capital \\
GM & EVs, legal and regulatory, market share, capacity utilization rate, charging infrastructure \\
\hline
\end{tabular}
\caption{Topics that are trending up over the considered time period. Refer figures \ref{fig:trending_up} for plots from NVDA and LCID.}
\label{tab:trending_up}
\end{table}

\begin{figure*}[htbp]
    \centering
    \includegraphics[width=\textwidth]{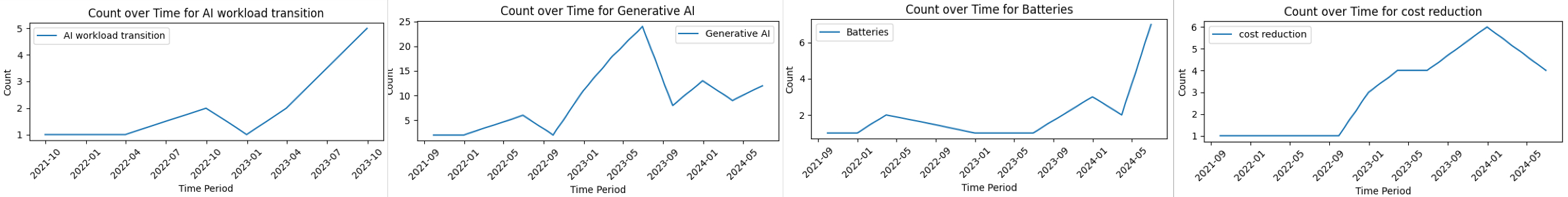}
    \caption{Trending up topics from NVDA (left) and LCID (right).}
    \label{fig:trending_up}
\end{figure*}

\subsection{Competitor Analysis}

Competitor analysis plays a crucial role in financial research by enabling analysts to benchmark a company's performance and strategic position relative to its industry peers. Traditionally, this comparison is performed using quantitative financial metrics such as revenue growth, sales volumes, profitability ratios, and valuation multiples (e.g., Price-to-Earnings ratios).

In this study, we propose a complementary methodology utilizing topics extracted from quarterly earnings calls as additional qualitative indicators of competitive advantage. Our approach involves identifying shared topics across competitors to highlight prevailing industry themes, thereby providing insights into sector-wide strategic movements. We calculate which 2 companies have been discussing similar non-product topics over the given time period. We sample top 100 most frequent topics for each company and compute Jaccard similarity over those topics. Tables \ref{tab:jaccard_similarity} and \ref{tab:jaccard_similarity_ev} provide an overview of how similar any two competitors in each sector are . As part of the equity analyst workflow, this can be further explored by exploring those individual topics (Table \ref{tab:nvda_common_topics} and Table \ref{tab:tsla_common_topics}) and noting the reason for similarity/differences through the extracted excerpts (Appendix).

Similarly, we can also analyze topics unique to each company within its industry sector, which can help identify competitive strengths and weaknesses. Such unique topics reveal strategic differentiators and potential advantages or weaknesses specific to individual companies, enriching traditional financial analyses with deeper strategic context and forward-looking perspectives.

\subsection{Emerging Topics}

Our last analysis in this work identifies emerging topics over the multi-year time-period. While topics can emerge on a rolling basis over the entire time period, in this analysis we look for topics that weren't mentioned in 2021 and 2022 but were prominent in the next 2 years over the entire dataset. Again, we filter by non-product topics and list some of the prominent ones that are aligned with the transformation in the semiconductor and EV industries:

\begin{tcolorbox}[colback=white, colframe=black, boxrule=0.4pt, arc=2pt, left=4pt, right=4pt, top=2pt, bottom=2pt, enhanced jigsaw]
\small
\textbf{Semiconductors:} Enterprise AI Platform, on-premises AI, AI copilots, mature semiconductor market, next-generation platform, high bandwidth memory, CPU market conditions, AI handsets, server demand.

\vspace{4pt}
\textbf{EV:} Low-cost vehicles, Hybrid vehicles, operating lease residual values, operational flexibility, capital efficiency, mid-decade profitability, future vehicle models, total addressable market.
\end{tcolorbox}

\begin{table*}[htbp]
\centering
\footnotesize
\begin{tabular}{|p{0.19\textwidth}|p{0.19\textwidth}|p{0.19\textwidth}|p{0.19\textwidth}|p{0.19\textwidth}|}
\hline
\textbf{AMD (19)} & \textbf{INTC (12)} & \textbf{ASML (8)} & \textbf{AVGO (16)} & \textbf{QCOM (8)} \\
\hline
AI inference, AI infrastructure, AI strategy, brand partnerships, buybacks, capital expenditures, cloud partnerships, data center, gaming, LLMs, supercomputing, data center growth, data center revenue, earnings call logistics, earnings release, edge computing, enterprise AI partnerships, gaming revenue, software ecosystem & AI infrastructure, AI strategy, AI workload transition, brand partnerships, capital expenditures, data center, dividends, GAAP financial measures, supply chain disruptions, trade war, data center growth, edge computing & AI computing demand, buybacks, capital expenditures, dividends, GAAP financial measures, supply chain disruptions, trade war, earnings call logistics & AI infrastructure, buybacks, capital expenditures, data center, dividends, Ethernet networking, GAAP financial measures, IT infrastructure demand, compute offload, data center infrastructure, earnings call logistics, earnings release, generative AI revenue growth, networking, networking revenue, software revenue & brand partnerships, buybacks, China re-opening, dividends, LLMs, earnings call logistics, edge computing \\
\hline
\end{tabular}
\caption{Common topics between NVDA and competitors among the top 100 most frequent topics for each company. We only show leaf node topics and the count is mentioned next to the company name.}
\label{tab:nvda_common_topics}
\end{table*}

\begin{table*}[htbp]
\centering
\footnotesize
\begin{tabular}{|p{0.19\textwidth}|p{0.19\textwidth}|p{0.19\textwidth}|p{0.19\textwidth}|p{0.19\textwidth}|}
\hline
\textbf{F (15)} & \textbf{GM (23)} & \textbf{RIVN (17)} & \textbf{LCID (12)} & \textbf{PSNY (9)} \\
\hline
 EV adoption, EV tax credits, earnings, energy transition, free cash flow, inflation, labor costs, management change, transportation costs, autonomous driving, battery manufacturing, commodity prices, manufacturing efficiency, software updates, vertical integration & Advanced materials, COVID-19, EV adoption, EV tax credits, earnings, free cash flow, IRA, inflation, labor costs, lithium, occupational health \& safety, store \& factory openings, supply chain disruptions, transportation costs, autonomous driving, battery cell production, battery manufacturing, capacity expansion, commodity prices, cost reduction initiatives, manufacturing efficiency, manufacturing ramp, production & Brand partnerships, EV adoption, earnings, gross margin, IRA, inflation, manufacturing expansion, operating and maintenance expense, store \& factory openings, supply chain disruptions, unplanned outages, cost reduction initiatives, manufacturing efficiency, manufacturing ramp, production, software updates, vertical integration & Brand partnerships, earnings, gross margin, management change, store \& factory openings, supply chain disruptions, cost reduction initiatives, manufacturing efficiency, manufacturing ramp, production, software updates, vertical integration & Brand partnerships, free cash flow, gross margin, management change, operating and maintenance expense, store \& factory openings, unplanned outages, manufacturing ramp \\
\hline
\end{tabular}
\caption{Common topics between TSLA and competitors among the top 100 most frequent topics for each company. We only show leaf node topics and the count is mentioned next to the company name.}
\label{tab:tsla_common_topics}
\end{table*}

\section{Limitations \& Future Work}
Despite the promising results and flexibility of our GenAI-driven topic ontology system, several limitations remain that open avenues for future research.

\textit{Subjectivity in Defining Financial Topics:} One of the fundamental challenges in evaluating topic retrieval systems in the financial domain lies in the lack of a universally agreed-upon definition of what constitutes a “useful financial topic.” Our approach defines financial topics as those relevant to an equity analyst assessing company performance. However, this definition is inherently subjective and may diverge from how other works in the literature conceptualize financial topics, complicating direct comparisons and benchmarking.

\textit{Lack of Gold-Standard Annotations:} There is currently a dearth of benchmark datasets with gold labels for emerging financial topics \cite{boutaleb2024bertrend}. This severely limits our ability to benchmark the Topic Retriever and Ontology against existing baseline. Developing such annotated datasets, perhaps via expert-annotated corpora or crowdsourcing methods guided by financial professionals, represents an important direction for future work.

\textit{Noisy Topics:} Despite semantic verification mechanisms, noise, such as misclassifications or vague topic boundaries, can persist. Errors such as incorrect topic-to-parent associations, alias misclassifications, and non-standardized topic names occasionally arise. Future work could explore ways to enrich LLM prompts with additional context—such as topic-specific excerpts to make ontology construction more robust. 

\textit{Human-in-the-Loop Remains Essential:} Although our system is designed to operate autonomously, human oversight is essential in downstream applications \cite{gandouet2024distilled, aggarwal2024large}. Analysts must review and validate the retrieved topics to ensure contextual relevance and usability in high-stakes financial decision-making.

In summary, addressing these limitations will be an essential step in making this system more reliable and robust in the financial domain.

\section{Conclusion}
In this paper, we introduced a novel GenAI-driven agentic framework to dynamically retrieve and organize financial topics from earnings call transcripts. 
We demonstrated the utility of our framework through use cases applicable to equity analysts, including trend analysis, and competitor benchmarking. While our study validates the efficacy of using LLM-agents for financial topic retrieval, we recognized several limitations. Addressing these limitations will be the focus of our future work. Overall, our agentic topic retrieval framework provides an effective approach for enhancing financial information retrieval systems, offering analysts detailed and practical insights into company strategies and sector trends.


\bibliographystyle{unsrt}
\bibliography{main.bib}

\clearpage
\onecolumn
\appendix
\section{Appendix}
\subsection{LDA topics}
\begin{figure*}[htbp]
    \centering
    \includegraphics[width=\linewidth, alt={LDA Topic Heatmap.}]{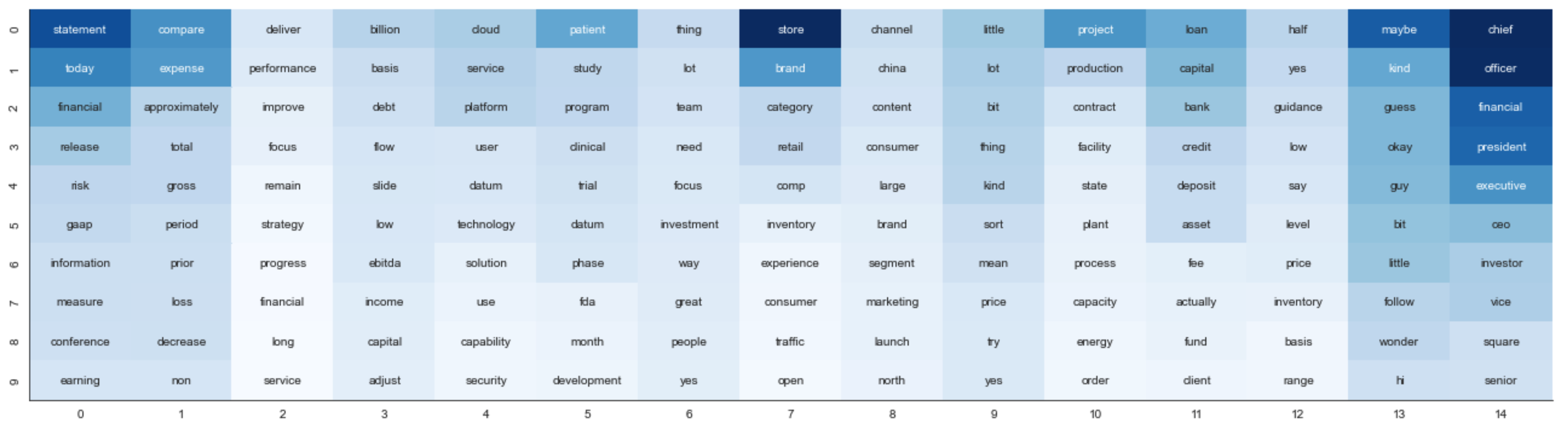}
    \caption{15 topics discovered by LDA baseline method represented by 10 keywords each where the color represents the u\_mass coherence score.}
    \label{fig:lda_heatmap}
\end{figure*}

\subsection{Prompt: Topic Retrieval}
\label{appendix:prompt_topic}
    \begin{lstlisting}
You are a financial analyst at a buy-side investment firm. Your broader goal is to extract insights on different topics by looking at a large corpus of financial documents. This will enable you to make financial decisions and investments of high-reward.
Your broader goal is broken down into different sub-tasks. The first sub-task here is as follows:
1. Extract topic from the given document text. These topics are of interest to a financial analyst. You are encouraged to tag multiple topics, both general and specific. 
1.a. Some examples of general topics: guidance, capex, dividends, buybacks, M&A, macro environment, labor, supply chain, pricing, consumer demand
1.b. Some examples of specific topics: Cybertaxi, Optimus, 48-volt architecture, A100, Red Sea attacks
1.c. If there are no topics of relevance, you should output an empty list
1.d. When identifying topics, consider, but not limited to:
1.d.a. Key industry terms, processes, or entities mentioned (e.g., "pharmacy benefits managers")
1.d.b. Specific initiatives, products, or strategies discussed
1.d.c. Regulatory or market forces impacting the business
1.e. Each topic should be an individual concept. Do not concatenate two concepts in the topic name, e.g., "Autopilot and AI"

2. Also extract the excerpts that mention those topics.
2.a. You may re-word the excerpt to scope it out better and provide a succinct summary of the context.
2.b. Do not speculate or provide opinions. Base all observations strictly on the information present in the documents.

The topics that you extract in this task will also be assigned parent topics later on -- to group together topics of same category, either from the topics you extracted or create a new topic title to name the category.

You final response should only be a JSON object like this, example:
[{"topic_name": "Guidance", "excerpts": ["excerpt 1", "excerpt 2"]}]

Your response should be parseable directly into JSON. Do not include any pre-text like 'Here is my response' etc.
\end{lstlisting}
\needspace{5\baselineskip}
\subsection{Prompt: Topic Existence}
\label{appendix:prompt_topic_exists}
\begin{lstlisting}
You are a semantic topic matcher. Your task is to determine if a given topic is semantically equivalent to any topics in a predefined list. The matches should be true semantic equivalents, not broader categories or parent topics that merely contain the query topic.

Key Matching Rules:
1. Topics must be semantically equivalent in scope and specificity
2. IMPORTANT: Matched topics must NOT be parent categories or supersets of the query topic
3. The relationship must be bidirectional - each topic should be able to substitute for the other

Examples of Invalid Matches (Parent/Subset Relationships):
 Query: "Brand and product design" != "Marketing and Advertising"
   Reason: Marketing is a broader category that contains branding, not an equivalent
 Query: "Product differentiation" != "Business Strategy"
   Reason: Business strategy encompasses product differentiation, making it a parent category
 Query: "Market expansion" != "Business Strategy"
   Reason: Market expansion is a subset of business strategy, not an equivalent concept

Examples of Valid Matches (True Semantic Equivalents):
 Query: "Customer acquisition" = "User growth"
   Reason: Both refer to the same concept at the same level of specificity
 Query: "Product analytics" = "Usage metrics analysis"
   Reason: Both describe the same activity with different terminology

Process each topic match request using these steps:
1. Analyze the core meaning and key concepts of the query topic
2. For each topic in the list, evaluate:
   - Core concepts and ideas
   - Subject matter and domain
   - Scope and specificity level
   - Intent and context
   - Parent/subset relationship check
3. For potential matches:
   - Verify its not a parent category or superset
   - Rate similarity on a scale of 0-100%
   - Explain your reasoning with explicit mention of scope equivalence

Given a List of reference topics and a Query topic, your response should be as a JSON object as follows:
<structured_output>
{
    "query_topic": string,
    "matches": [
        {
            "topic": string,
            "similarity": number
        }
    ],
    "detailed_analysis": {
        "matched_topics": [
            {
                "topic": string,
                "similarity": number,
                "reasoning": string,
                "parent_subset_check": string
            }
        ]
    }
}
</structured_output>

Example input:
Reference topics:
- Digital marketing strategy
- Online customer acquisition
- User growth techniques
- Social media advertising
- Business development

Query topic: Web user acquisition

Example output:
<structured_output>
{
    "query_topic": "Web user acquisition",
    "matches": [
        {
            "topic": "Online customer acquisition",
            "similarity": 95
        },
        {
            "topic": "User growth techniques",
            "similarity": 90
        }
    ],
    "detailed_analysis": {
        "matched_topics": [
            {
                "topic": "Online customer acquisition",
                "similarity": 95,
                "reasoning": "Both terms describe the specific process of gaining new users through online channels",
                "parent_subset_check": "Neither term is broader than the other; they describe the same scope of activity"
            },
            {
                "topic": "User growth techniques",
                "similarity": 90,
                "reasoning": "Both terms refer to the same core activity of expanding user base",
                "parent_subset_check": "Both terms operate at the same level of specificity, focusing on the tactical aspect of gaining users"
            }
        ]
    }
}
</structured_output>
\end{lstlisting}

\subsection{Prompt: Topic Insertion}
\label{appendix:prompt_topic_insert}
\begin{lstlisting}
You are an expert ontologist agent working on a financial information retrieval system. Your task is to assign a financial **topic** to the most appropriate parent topic in an ontology tree. The parent topic should be:

- Semantically relevant to **topic** (the topic should logically belong under this parent)
- You may only assign a parent that is another topic from the tree.
- You must aim for **maximum specificity**. If two parents are possible, prefer the more granular one. The most specific appropriate parent available (avoid assigning to very broad categories like "Artificial Intelligence" when more specific parents like "Machine Learning" exist)
- Do not skip levels.
- Do not invent new categories or topics.

The topic tree ontology is structured as follows: JSON object of {"super-parent topic": [list of child topics...]..}

Planning:
1. Start by finding the most appropriate super-parent topic for the given topic
2. Identify the most specific child topic that can be a parent of the give **topic**
3. If there is no such parent present, then assign the super-parent topic as the parent topic of the given **topic**


Input:
- Given Topic: [Name of the topic to categorize]
- Topic Tree: {Super-parent topic: [list of potential parent topics], super-parent topic 2: [list of potential parent topics]...}

Output:
- Parent Topic: [The most appropriate parent topic from the list]
- Reasoning: [Brief explanation of why this parent was selected]

Example 1:
Input:
- Given Topic: "Roboadvisor"
- Topic Tree: {'Technology and Innovation': ['5G', 'Automation', 'Batteries'], 'Environmental Issues': ['Air Quality', 'Biodiversity', 'Carbon Neutral'], 'Financial Technology': ['Digital Payments', 'Digital Wallet', 'Fintech']}

Output:
<structured_output>
{    
    "reasoning": "Roboadvisor can be assigned to super-parent Financial Technology, and within that I found Fintech as the most granular parent topic.",
    "parent": "Fintech"
}

Example 2:
Input:
- Given Topic: "Robotics"
- Topic Tree: {'Technology and Innovation': ['5G', 'Automation', 'Batteries'], 'Environmental Issues': ['Air Quality', 'Biodiversity', 'Carbon Neutral'], 'Financial Technology': ['Digital Payments', 'Digital Wallet', 'Fintech']}

Output:
<structured_output>
{    
    "reasoning": "Robotics can be assigned to super-parent Technology and Innovation, and within that I did not find any more granular parents for Robotics.",
    "parent": "Technology and Innovation"
}

</structured_output>
Verify that the parent topic in output exists in the list of Potential Parent Topics.
\end{lstlisting}

\section{Disclaimer}

This research explores generative AI techniques for topic extraction and ontology construction from earnings calls. This work is not related to the AI system used in Bloomberg's \textit{AI-Powered Earnings Call Summaries (https://www.bloomberg.com/company/press/bloomberg-launches-ai-powered-earnings-call-summaries/)} solution.
\end{document}